# The Tower of Babel Meets Web 2.0: User-Generated Content and Its Applications in a Multilingual Context

Brent Hecht[*] and Darren Gergle[*†]
Northwestern University
[*]Dept. of Electrical Engineering and Computer Science, [†] Dept. of Communication Studies
brent@u.northwestern.edu, dgergle@northwestern.edu

## ABSTRACT
This study explores language's fragmenting effect on user-generated content by examining the diversity of knowledge representations across 25 different Wikipedia language editions. This diversity is measured at two levels: the concepts that are included in each edition and the ways in which these concepts are described. We demonstrate that the diversity present is greater than has been presumed in the literature and has a significant influence on applications that use Wikipedia as a source of world knowledge. We close by explicating how knowledge diversity can be beneficially leveraged to create "culturally-aware applications" and "hyperlingual applications".

**ACM Classification Keywords**
H.5.3. [Information Interfaces and Presentation]: Group and Organization Interfaces – collaborative computing, computer-supported cooperative work

**General Terms**
Human Factors

## INTRODUCTION
A founding principle of Wikipedia was to encourage consensus around a single neutral point of view [18]. For instance, its creators did not want U.S. Democrats and U.S. Republicans to have separate pages on concepts like "Barack Obama". However, this single-page principle broke down in the face of one daunting obstacle: language.

Language has recently been described as "the biggest barrier to intercultural collaboration" [29], and facilitating consensus formation across speakers of all the world's languages is, of course, a monumental hurdle. Consensus building around a single neutral point of view has been fractured as a result of the Wikipedia Foundation setting up over 250 separate language editions as of this writing. It is the goal of this research to illustrate the splintering effect of this "Web 2.0 Tower of Babel"[1] and to explicate the positive and negative implications for HCI and AI-based applications that interact with or use Wikipedia data.

We begin by suggesting that current technologies and applications that rely upon Wikipedia data structures implicitly or explicitly espouse a *global consensus hypothesis* with respect to the world's encyclopedic knowledge. In other words, they make the assumption that encyclopedic world knowledge is largely consistent across cultures and languages. To the social scientist this notion will undoubtedly seem problematic, as centuries of work have demonstrated the critical role culture and context play in establishing knowledge diversity (although no work has yet measured this effect in Web 2.0 user-generated content (UGC) on a large scale). Yet in practice, many of the technologies and applications that rely upon Wikipedia data structures adopt this view. In doing so, they make many incorrect assumptions and miss out on numerous technological design opportunities.

To demonstrate the pitfalls of the global consensus hypothesis – and to provide the first large-scale census of the effect of language in UGC repositories – we present a novel methodology for assessing the degree of world knowledge diversity across 25 different Wikipedia language editions. Our empirical results suggest that the common encyclopedic core is a minuscule number of concepts (around one-tenth of one percent) and that sub-conceptual knowledge diversity is much greater than one might initially think—drawing a stark contrast with the global consensus hypothesis.

In the latter half of this paper, we show how this knowledge diversity can affect core technologies such as information retrieval systems that rely upon Wikipedia-based semantic relatedness measures. We do this by demonstrating knowledge diversity's influence on the well-known technology of Explicit Semantic Analysis (ESA).

In this paper, our contribution is four-fold. First, we show that the quantity of the world knowledge diversity in Wikipedia is much greater than has been assumed in the literature. Second, we demonstrate the effect this diversity can have on important technologies that use Wikipedia as a source of world knowledge. Third, this work is the first large-scale and large-number-of-language study to describe some of the effects of language on user-generated content. Finally, we conclude by describing how a more realistic perspective on knowledge diversity can open up a whole new area of applications: *culturally aware applications*, and its important sub-area, *hyperlingual applications*.

---
[1] See [18] for original use of the analogy.

## BACKGROUND

Wikipedia has in recent years earned a prominent place in the literature of HCI, CSCW, and AI. It has served as a laboratory for understanding many aspects of collaborative work [6, 16, 17, 23, 26] and has also become a game-changing source for encyclopedic world knowledge in many AI research projects [11, 20, 28, 30]. Yet the vast majority of this work has focused on a single language edition of Wikipedia, nearly always English. Only recently has the multilingual character of Wikipedia begun to be leveraged in the research community, ushering in a potential second wave of Wikipedia-related research.

This new work has been hailed as having great potential for solving language- and culture-related HCI problems [14]. Unfortunately, while pioneering, most of this research has proceeded without a full understanding of the multilingual nature of Wikipedia. In the applied area, Adafre and de Rijke [1] developed a system to find similar sentences between the English and Dutch Wikipedias. Potthast and colleagues [25] extended ESA [11, 12] for multilingual information retrieval. Hassan and Mihalcea [13] used an ESA-based method to calculate semantic relatedness measurements between terms in two different languages. Adar and colleagues [2] built Ziggurat, a system that propagates structured information from a Wikipedia in one language into that of another in a process they call "information arbitrage". A few papers [22, 27] attempted to add more interlanguage links between Wikipedias, a topic we cover in detail below.

### The Global Consensus of World Knowledge

Though innovative, much of the previous work implicitly or even explicitly adopts a position consistent with the global consensus hypothesis. The global consensus hypothesis posits that every language's encyclopedic world knowledge representation should cover roughly the same set of concepts, and do so in nearly the same way. Any large differences between language editions are treated as bugs that need to be fixed, discrepancies that will go away in time (and possibly need help doing so), or a problem that should simply be ignored. For example, Sorg and Cimiano [27] state that since English does not cover a large majority of German's concepts, there is "clearly" something wrong with the links.

"Information arbitrage", and applications designed to support it [2], provides an especially interesting case. Innate to the approach is the assumption that the information in one language edition is inherently useful to any other language edition that is missing it, which, if not qualified, is a strong case of the global consensus hypothesis. We argue that although this "information arbitrage" model in many cases provides risk-free informational profit, it does not always do so. In fact, if such an approach is applied heedlessly it runs the risk of tempering cultural diversity in knowledge representations or introducing culturally irrelevant/culturally false information. Our results can be used in tandem with approaches such as information arbitrage to provide a framework for separating "helpful" arbitrage, which can have huge benefits, from "injurious" arbitrage, which has negative qualities.

### Evidence Against the Global Consensus Hypothesis

While fewer in number, some recent papers have focused on studying the *differences* between language editions. These papers fall more in line with social science positions regarding the nature of knowledge diversity. Our prior work [14] demonstrated that each language edition of Wikipedia focuses content on the geographic culture hearth of its language, a phenomenon called the "self-focus" of user-generated content. In a similar vein, Callahan and Herring [7] examined a sample of articles on famous persons in the English and Polish Wikipedia and found that cultural differences are evident in the content.

## OUR APPROACH

In order to question the veracity of the global consensus hypothesis, quantify the degree of knowledge diversity, and demonstrate the importance of this knowledge diversity, a large number of detailed analyses are necessary. The following provides an overview of the various stages of our study.

**Step One** ("Data Preparation and Concept Alignment"): First, we develop a methodology similar to previous literature [2, 13, 25] that allows us to study knowledge diversity. This methodology – highlighted by a simple algorithm we call CONCEPTUALIGN – aligns concepts across different language editions, allowing us to formally understand that "World War II" (English), "*Zweiter Weltkrieg*" (German), and "*Andre verdenskrig*" (Norwegian) all describe the same concept. Since the methodology leverages user-contributed information, we also provide an evaluation to demonstrate its effectiveness.

**Step Two** ("World Knowledge Diversity in Wikipedia"): After describing the concept alignment process, we can analyze the extent of diversity of knowledge representations in different language editions of Wikipedia. These analyses take place at both the conceptual (article) level and the sub-conceptual (content) level, and serve two purposes: (1) they show that the global consensus hypothesis is false, and (2) they provide the first empirically derived quantitative descriptions of the extent of knowledge diversity across numerous Wikipedia language editions.

**Step Three** ("The Effect of Diversity on Technologies"): Next, we provide evidence that knowledge diversity across language editions actually affects important technologies that use Wikipedia as a source of world knowledge. Our test case is Explicit Semantic Analysis [11], a technology that has been widely applied in fields ranging from HCI to AI to NLP.

**Step Four** ("Implications for Design"): Finally, we conclude with a discussion of the implications for design that result from each of the previous analyses. We highlight the importance of considering language and culture in application building, and introduce two ideas: *culturally-aware applications* and *hyperlingual applications*.

## DATA PREPARATION AND CONCEPT ALIGNMENT

A large data preparation process preceded the studies described in this paper, all of which were executed using an extension of our open-source WikAPIdia software. WikAPIdia is a MySQL-based Java API to the database dumps made available by the Wikimedia Foundation. While support for any existing language edition can easily be added to WikAPIdia, our current implementation uses the 25 different language editions in Table 1 (the 25 largest language editions in Wikipedia as of mid-2009[2], when the database dumps were downloaded).

### Parsing

For each language edition database dump, WikAPIdia parses out article metadata (title, id, etc.), links from one article to another, interlanguage links between articles of different language editions, disambiguation pages (polysemy), and redirects (synonymy).

| Language | # Articles | # Outlinks |
|---|---|---|
| Catalan | 191,770 | 5,354,217 |
| Chinese | 257,257 | 7,100,654 |
| Dutch | 552,783 | 11,251,251 |
| English (Largest in Study) | 3,003,868 | 91,133,413 |
| French | 839,774 | 25,964,178 |
| German | 956,571 | 25,663,974 |
| Hebrew (Smallest in Study) | 96,197 | 3,498,739 |
| Italian | 607,890 | 18,224,304 |
| Japanese | 612,450 | 25,461,189 |
| Norwegian | 225,370 | 4,589,390 |
| Russian | 420,256 | 10,813,981 |
| Spanish | 509,145 | 14,974,843 |
| … | … | … |
| TOTAL | 11,066,552 | 304,030,822 |

**Table 1** – A brief overview of the size of some of the 25 language editions in our study. Other languages included are: Czech, Danish, Finnish, Hungarian, Indonesian, Korean, Polish, Portuguese, Romanian, Slovak, Swedish, Turkish, and Ukrainian. The median number of articles was 225,370.

### Interlanguage Links and the CONCEPTUALIGN Algorithm

The basis for our concept alignment process is the enormous number of *interlanguage links* (ILLs) manually entered or propagated by bots in Wikipedia. ILLs serve to link an article in one language edition to an article on the same concept in another language edition. For example, an ILL exists that links the English article "Computer Science" to the Catalan article on the same topic, "*Informàtica*". On the live Wikipedia site, ILLs included in an article can be viewed and clicked on under the "languages" (or "*en otros idiomas*", "*In anderen Sprachen*", etc.) heading at the bottom of the left sidebar that accompanies each article.

In the 25 languages used in our study, we parsed out around 52 million separate ILLs. These links are partially generated by a group of bots that can, for example, add a German → English ILL when it finds that an English → German ILL has been manually added. However, it is crucial to our study that we have an accurate collection of interlanguage links in order to ascertain the degree of knowledge diversity present. While the bots propagate these links across the various Wikipedia language editions, their lack of formality means that we cannot be sure that they are exhaustive. To address this lack of formality and help maximize the information value from the user-generated ILLs, we implemented an algorithm, CONCEPTUALIGN, which formally categorizes every article in every language into a single concept group.

ILLs are typically viewed as pairwise dictionary-like entities, e.g. as done in [9]. However, the ILL dataset can also be viewed as a set of directional edges (ILLs) and nodes (articles) that form an enormous number of individual connected components in a graph containing all the articles in our study (see Figure 1). Each connected component represents a concept that is described in the articles in the component. CONCEPTUALIGN was designed for this graph-based view of interlanguage links. The algorithm picks a node at random and does a breadth-first search, ignoring edge direction, until it finds all connected nodes in the component. It then labels all those nodes as belonging to the same concept. Next, it picks another node (from a different connected component), and continues the process until all nodes have been labeled with a concept.

CONCEPTUALIGN effectively adds to the ILL dataset new ILLs that make each separate component fully connected; all articles in a concept group are connected to all other articles in the group with an ILL following the operation of the algorithm. In this way, CONCEPTUALIGN maximizes the information content derived from the user-generated ILL dataset. In other words, there need not be a large number of Wikipedians proficient in both Korean and Slovak. CONCEPTUALIGN only requires is that there be a large number of Wikipedians proficient in Korean and English, and Slovak and English (for example). This is an approach much more suited to the real-life patterns of bilingualism.

### Evaluation of CONCEPTUALIGN

Like all Wikipedia data used in research, ILLs are subject to concerns regarding the quality of user-generated content. The concern in the context of this work is that of the "missing ILL". A missing ILL could prevent CONCEPTUALIGN from correctly "merging" two conceptually identical disconnected components in cases where no other ILL serves the same role. Therefore, we evaluate in two different ways the combined effectiveness of the ILL dataset and CONCEPTUALIGN. The first approach is based on the algorithm's nature as a "missing link finder" and places it in the context of state-of-the-art machine learning techniques that have been developed to accomplish the same task [22, 27]. In the second evaluation, we examine the effectiveness of the algorithm using bilingual human coders.

---

[2] The Arabic Wikipedia would have been included, but it has a few alternative encoding norms that would have invalidated some of our results.

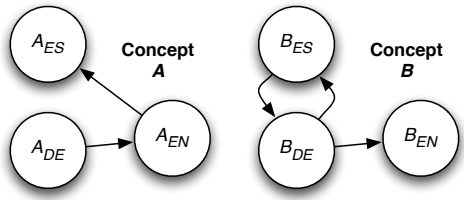

**Figure 1: A simplified ILL graph. Interlanguage links are represented as edges, and articles as nodes. For example, the article on concept *A* in English ($A_{EN}$) has an interlanguage link to the article on concept *A* in Spanish ($A_{ES}$). CONCEPTUALIGN simply finds all the nodes (articles) in a connected component (e.g. *A*), identifies them all as belonging to a single concept, and moves to the next component (e.g. *B*). Importantly, note that even though $B_{EN}$ and $B_{ES}$ have zero interlanguage links to each other, they are still identified as part of the same concept. The same is true for $A_{ES}$ and $A_{DE}$.**

**Algorithm Evaluation Study:** The first step in examining CONCEPTUALIGN's effectiveness in finding missing ILLs is to compare its performance to existing state of the art techniques in missing ILL detection. Sorg and Cimiano [27] published a dataset of the missing German → English ILLs found by their machine learning approach that makes use of a support vector machine (SVM) on a subset of German articles. CONCEPTUALIGN found ILLs for 95.8% of the German/English "missing ILLs" in the dataset. Interestingly, 91.8% of these links had been added manually into Wikipedia since the time of Sorg and Cimiano's study. The remaining 4.0% of ILLs were discovered by CONCEPTUALIGN.

The precision of the SVM that generated the missing links is far from perfect[3] and some found links are to articles not included in our study (Wikipedia project pages, etc). This leads us to conclude that CONCEPTUALIGN (and its underlying data) is *at least* as effective as the SVM. We write "at least" because we have no idea how many additional missing links CONCEPTUALIGN would have found on the same sample of articles on which Sorg and Cimiano tested their SVM.

---

[3] Exact precision figures are not available due to the fact that 7.6% of the dataset was not comparable due to changes in Wikipedia article structure and Sorg and Cimiano's inclusion of Wikipedia project pages (which were not included in our evaluation).

**Human Evaluation Study:** Despite the effectiveness of CONCEPTUALIGN relative to the state of the art in missing ILL detection, in order to determine the degree of world knowledge diversity across Wikipedia language editions it is important to get an accurate estimate of the number of ILLs that are likely missing *after running the algorithm.* While it would be impractical to evaluate the coverage of our dataset for all 25*24 = 600 language pairs in our study by hand, we did perform this evaluation on three pairs of languages: English paired with Spanish, Japanese, and Italian. Large-Wikipedia languages were used because they have the biggest impact on the results we present in the latter part of this paper. Moreover, due to patterns of bilingualism around the world, these languages are most likely to play an important role in connecting concept components in the ILL graph.

After CONCEPUALIGN was run on our entire dataset, six human coders (two for each language) were recruited, each highly proficient in both languages $L_1$ and $L_2$. Each coder was given 75 primary Monolingual ($L_1|L_2$) articles and 75 secondary Monolingual($L_2|L_1$) articles, as identified by CONCEPTUALIGN. Monolingual($L_X|L_Y$) articles are $L_x$ articles representing concepts that CONCEPTUALIGN has indicated do not have articles in $L_Y$. For each article, coders were instructed to attempt to find a conceptual equivalent in the other language of the language pair. They were told to use any tool or information they wished to accomplish this task, but were limited to five minutes per article. Coders were also given a list of 25 Bilingual($L_1, L_2$) articles to evaluate the precision of CONCEPTUALIGN (and the ILL graph as a whole). For these articles, coders were instructed to determine whether or not each bilingual article pair covered the same concept. For each language pair, we used the union of the coders' results. Any disagreements between coders were resolved by discussion. Finally, coders were told to ignore any disambiguation pages in the sample.

| Language Pair | P(missing link) | P(incorrect link) |
|---|---|---|
| Italian / English | 0.08 | 0.00 (none) |
| Japanese / English | 0.02 | 0.00 (none) |
| Spanish / English | 0.06 | 0.00 (none) |

**Table 2: Results from our evaluation of CONCEPTUALIGN and the interlanguage links that power it.**

As shown in Table 2, the probability of a missing link is low, especially in the case of Japanese/English. In addition, the precision of the ILL graph appears to be good, as not a single incorrect link was found. The results from both evaluations together provide evidence supporting the relatively high quality of concept alignment, and also provide us with a reasonable estimate of the amount of error we are likely to find as a result of missing ILLs.

### WORLD KNOWLEDGE DIVERSITY IN WIKIPEDIA

In this section of the paper, we demonstrate the extensive diversity of world knowledge found in our 25-language dataset. First, we examine the global consensus hypothesis at a concept level. The notion underlying the hypothesis in this context is that every language's encyclopedic world knowledge representation should cover roughly the same set of concepts. Given the differential in article quantity in Table 1, this assumption is most often instantiated in the corollary that the English Wikipedia, given its massive size, is a near-complete superset of all the other Wikipedias. Every other language edition thus is supposed to cover some subset of the English editions' articles.

Second, we explore the global consensus hypothesis at the sub-conceptual level. Here the hypothesis posits that two articles about the same concept in two different languages will describe that concept roughly identically. For instance, all articles on "Psychology" would describe "Psychology" in close to exactly the same fashion.

Before continuing with our analyses, it is important to briefly discuss in more detail the two information levels investigated. At the higher conceptual level, the topic of the article matters and the way that topic is defined is unimportant. The exploration of the global hypothesis at the sub-conceptual level is predicated on the higher-level assumptions being true.

**Concept Diversity**

*Method*
Conceptual diversity is measured by the degree of concept coverage overlap across different language editions of Wikipedia. Concept coverage overlap is determined as follows: if one or more articles exist on a concept $C$ in a language $L_1$'s Wikipedia, then $C$ can be considered to be covered by $L_1$'s Wikipedia world knowledge representation. As such, if an article on $C$ exists in both languages $L_1$ and $L_2$, $C$ can be considered to be covered by the shared world knowledge of $L_1$ and $L_2$. One can examine the conceptual overlap between $n$ languages using the same method, but including only concepts that are covered by the knowledge intersection of all $n$ languages. For instance, we can approximate shared "global knowledge", or the "encyclopedic core", with $n = 25$.

*Results*
Our results (Figure 2, Table 3) demonstrate that a surprisingly small amount of concept overlap exists between languages of Wikipedia, refuting the global consensus assumption at the concept level. Over 74 percent of concepts are described in only one language. Moreover, more than 95.5 percent of concepts appear in six or fewer languages, indicating that even if we were to cut out some of the less developed Wikipedias, poor concept overlap – and thus great conceptual diversity – would still exist.

To further support this last point, we ran the same experiment using only the three largest Wikipedias (English, German, and French) and the largest six Wikipedias (all of which have 600,000 or more articles). In the first case, around 80 percent of concepts remained single-language, while only 7 percent were in all three. In the second case, 77 percent of concepts were single-language, while only 1.5 percent are in were in all six.

Our 25-language analysis also revealed the extremely small number of concepts that are covered in all 25 language editions: 6,966 concepts (0.12 percent of all concepts). While small in number, these concepts are revealing and could be considered "globally relevant". Table 4 lists a few of these concepts and their corresponding articles in English, Japanese, Slovak, and Dutch.

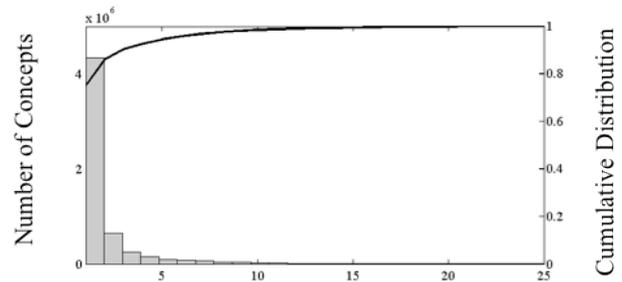

**Figure 2: The amount of concept overlap in our 25-language Wikipedia data set. The CDF (black line) represents the cumulative proportion of concepts that are shared by the number of languages on the x-axis.**

It is important to note that we have not yet taken into account the small number of missing ILLs that may remain after CONCEPTUALIGN is run. While it is impossible to predict exactly how missing ILLs would effect the distribution in Figure 2, our earlier studies allow us to establish a realistic range. A conservative estimate would be to assume a "best case scenario" for the global consensus hypothesis: that all of the missing ILLs would link single-language concepts to articles in other languages, that no missing ILL is a reflexive "duplicate" of another, and that the Italian performance rate (the worst we found) occurs across the entire dataset. Even in this "best case scenario" for the global consensus assumption, 68.2 percent of concepts remain unique to a single language. This still represents a great deal of conceptual diversity.

**Table 3: Pairwise conceptual coverage overlap. Each cell represents the ratio of concepts in the column's language edition covered by the row's language edition.**

|  | *Cata* | *Chinese* | *Czech* | *Dan* | *Dutch* | *Engl* | *Finn* | *Fren* | *German* | *Hebr* | *Hun* | *Indo* | *Ital* |
|---|---|---|---|---|---|---|---|---|---|---|---|---|---|
| **German** | 0.38 | 0.36 | 0.57 | 0.55 | 0.41 | 0.16 | 0.53 | 0.35 |  | 0.58 | 0.48 | 0.43 | 0.38 |
| **English** | 0.58 | 0.56 | 0.66 | 0.66 | 0.65 |  | 0.70 | 0.60 | 0.51 | 0.75 | 0.66 | 0.57 | 0.65 |
| **French** | 0.45 | 0.36 | 0.51 | 0.51 | 0.43 | 0.17 | 0.52 |  | 0.31 | 0.58 | 0.48 | 0.45 | 0.44 |
|  | *Japan.* | *Korean* | *Norw* | *Pol* | *Portu* | *Rom* | *Russ* | *Slov* | *Spanish* | *Swed* | *Turk* | *Ukr* |  |
| **German** | 0.24 | 0.45 | 0.46 | 0.34 | 0.33 | 0.46 | 0.39 | 0.48 | 0.36 | 0.46 | 0.39 | 0.32 |  |
| **English** | 0.41 | 0.61 | 0.64 | 0.62 | 0.66 | 0.64 | 0.57 | 0.59 | 0.63 | 0.62 | 0.51 | 0.40 |  |
| **French** | 0.25 | 0.45 | 0.44 | 0.36 | 0.41 | 0.37 | 0.36 | 0.38 | 0.43 | 0.43 | 0.38 | 0.32 |  |

In addition to examining the global properties of the concept overlap among all languages, we can perform a more detailed examination of the pairwise diversity across languages. As shown in Table 3, the English-As-Superset corollary to the global consensus hypothesis does not hold. Despite its massively larger size, the English Wikipedia (row 2) covers no more than approximately three-quarters of any other Wikipedia in our study. The case of overlap between German and English – two very mature language editions – is quite illustrative. English is more than three times the size of German, but only covers slightly more than 50 percent of its concepts.

| English | Japanese | Slovak | Dutch |
|---|---|---|---|
| Auguste Rodin | オーギュスト・ロダン | Auguste Rodin | Auguste Rodin |
| Sarah Palin | サラ・ペイリン | Sarah Palinová | Sarah Palin |
| Turing Machine | チューリングマシン | Turingov stroj | Turing maschine |
| August 25 | 8月25日 | 25. august | 25 augustus |
| Chelsea F.C. | チェルシーFC | Chelsea FC | Chelsea FC |

**Table 4: Example "global" concepts (*n* = 25). Others include "Britney Spears", "Periodic Table", and "Milk".**

At first glance, many of these statistics may appear rather remarkable. However, any dedicated Wikipedian could provide ample anecdotal support. For instance, the entry on *Prinzipalmarkt*, a key commercial district in the mid-sized city of Münster, Germany, remains German-only, despite the district's local significance. Similarly, the American country music duo Big and Rich, who have sold millions of records in the United States, have an English-only article.

Even seemingly "major" concepts in a culture, say minor league baseball, are not pervasively included across all Wikipedias in our study. In fact, only 40 percent of languages in our study have an article on minor league baseball. Of course, to the global sports fan, this makes sense. Why would the Poles, Finns, Russians, etc. spend their time writing about minor league baseball when most of them do not even care about the sport of baseball in general? (Of course there will be individual exceptions.) The overlap gets even smaller when one takes a step down the specificity hierarchy and considers minor league baseball teams. For instance, the Lansing Lugnuts, a Class A minor league team, have an English-only article.

*Important Sidenote: Semantic Drift*
In a small number of cases, more than one article per language belongs to a single concept. If prominent across our dataset, this could have unpredictable effects on our results. We measured the average number of languages per article in a concept (a connected component of the ILL graph), a ratio we call conceptual *clarity*. A concept with clarity equal to 1.0 indicates *perfect clarity*, or exactly one article per language in the concept. Fortunately, perfect clarity is by far the norm in our dataset, with only rare instances of very low clarity. In fact, only about 2,700 concepts in total have a clarity of less than 0.5, which can still be reasonable in some cases. These statistics are bolstered by the precision results from our coders. Further analysis of the rare cases of low clarity can be found in [4].

**Sub-Concept Diversity**
Even when two language editions cover the same concept (with perfect clarity), they may describe that concept differently. If a significant phenomenon, this *sub-concept* diversity would add extensively to the overall diversity of world knowledge representations present in Wikipedia. In this sub-section, we show that sub-concept diversity is a prominent force, although similarities do exist on average between articles on the same topic. As a result, while false overall, the global consensus hypothesis does have some truth at the sub-concept scale.

*Method*
Our experiment on sub-concept diversity borrows from [1] the idea of using outlinks, or links in one article pointing to another article, as a "highly focused entity-based representation of [natural language]." In other words, outlinks[4] provide a decent structured, canonical/ language-independent summary of raw text.

Operating under this assumption, we compared the outlinks of each of the "global concepts" to determine the degree to which the articles covered the same content. If two articles on the same concept (e.g. "Sarah Palin") in two languages define the concept in a nearly identical fashion, they should link to articles on nearly all the same concepts (e.g. "John McCain", "Fox News", etc.). If, on the other hand, there is great sub-concept diversity, these articles would link to very few articles about the same concepts. If links in two different languages are pointed at the same concept, the destination of both links would be articles belonging to the same ILL graph component.

Our metric in this experiment was the Overlap Coefficient (*OC*), first used in the UGC domain in [21] and calculated as follows:

$$OC(C_{L1}, C_{L2}) = \frac{|Outlinks(C_{L1}) \cap Outlinks(C_{L2})|}{\min(|Outlinks(C_{L1})|, |Outlinks(C_{L2})|)}$$

where $C_{L1}$ and $C_{L2}$ are articles about concept *C* in languages $L_1$ and $L_2$, respectively.

*OC* is the size of intersection of the two sets of links divided by the size of the *smaller* of the two sets. In other words, *OC* describes the percentage of the links of the shorter of the Wikipedia articles about concept *C* also contained in the longer of the articles on *C*. *OC* is an ideal metric as it controls for the systematic differences in article length and some of the systematic differences in linking behavior. In this way, *OC* provides the "best case" scenario for the global consensus assumption in terms of these systematic differences.

---

[4] Outlinks are not the same as interlanguage links. Outlinks link to articles within the same language edition and appear as blue hyperlinks in the live version of Wikipedia.

Calculation of the *OC* was straightforward with one exception relating to links to time-related concepts. Because the norm about linking to these concepts is different in each of the Wikipedias studied, we used WikAPIdia's spatiotemporal package to filter out all links to years, dates, and months in our analyses. In languages where time-related links occur, these form a substantial percentage of outlinks on many pages, creating a signal that needed to be neutralized in order to explore more general sub-concept diversity.

For this study, we used a sample of concepts in the global concepts list, filtered for perfect clarity. We also required that each article have at least three outlinks, in order to make our experiment non-trivial, and three inlinks, to ensure that each article was relatively integrated into its Wikipedia. For each concept, we calculated the *OC* for every $L_1$, $L_2$ pair, leading to 600 pairs per concept. Our final sample consisted of over 217,000 of these pairs.

*Results*

The mean overlap coefficient for our sample was only 0.41 (SD=0.2). This means that, on average, the longer of two articles on the same concept contains only 41 percent of the outlinks in the shorter of the articles[5]. Adding in the assumption in [1], a longer article on a concept *C* only covers 41 percent of the content of a shorter article on *C*.

While the main driving force behind the concept-level diversity seemed to be cultural ("self-focus" in particular), the causes in the sub-concept context, on the other hand, seem to be more mixed in nature. Certainly, cultural forces are very prominent. In the case of the concept that is called "Psychology" in English, for example, the Spanish article ("*Psicología*") contains many outlinks to Latin American countries not contained in the German article ("*Psychologie*"). These links come from a section in the "*Psicología*" page about Latin America's contribution to psychology. While many of these culture-specific instances of sub-concept diversity are evident in links to articles on geographic entities (as is suggested in the results of [14]), this is by no means always the case. The article on "*Paz*" (Peace) in Spanish contains links to Christian concepts in a discussion on peace in the Bible, links that are not in the English Wikipedia's article on "Peace".

However, two other factors also seem to be at play. The first is linking behavior. There are some cases in which articles in two languages describe the exact same content, but one contains a hyperlink and the other does not. For instance, in the example of "Psychology", the Spanish article links to "*Biología*" (Biology) but the German one does not, even though both discuss biology. The last major factor is seemingly random differences in descriptions. While some of these could be less obviously or indirectly cultural differences, it is not unreasonable to think that two people of the same culture with access to the same information would describe a concept differently. Examining this phenomenon represents an important area of future work.

## THE EFFECT OF DIVERSITY ON TECHNOLOGIES

In this section, we explore the effect that the large diversity of representations established above has on technologies that use Wikipedia as a source of world knowledge. As a case study, we use one of the most generally applicable and popular technologies developed around Wikipedia, the semantic relatedness measure Explicit Semantic Analysis (ESA). ESA was first introduced by Gabrilovich and Markovitch [11], and has been shown to mimic human judgments on standard datasets better than any other semantic relatedness (SR) measure.

The semantic relatedness between two concepts can be defined as some measure of the number and strength of relationships between the concepts. Semantic similarity, perhaps more familiar to some readers, is a subset of semantic relatedness in which the only relations considered are hypernymy/hyponymy [5]. In addition to their implicit import as a model of human judgments, semantic relatedness measures play an essential role in a variety of technologies including information visualization [3, 15], information retrieval, word sense disambiguation, text summarization and annotation, determining the structure of texts, and lexical selection [5].

Our goal here is to determine if the diversity in world knowledge representations described in the previous study causes significantly different ESA scores for any two concepts $C_1$ and $C_2$. This is an effect that could greatly influence end-user applications (e.g. [3]).

## Method

We built an ESA implementation based on the descriptions in [8, 11, 25]. ESA models concepts as vectors of their abstract (bag-of-words-based) relationships with a set of other concepts. These "other concepts" are ESA's world knowledge representation, and are defined using Wikipedia articles. ESA compares the vectors of two input concepts $C_1$ and $C_2$ and the more similar the vectors, the higher the ESA value (SR). A full description can be found in [11, 12].

In implementing ESA for a large number of languages, we were forced to make several changes to the original implementation. Due to the unavailability of standardized tools for certain languages in this study (e.g. a Snowball stemmer, a stop word list), we could only make fair comparisons between ESA implementations based on the following ten languages: Spanish, Hungarian, Norwegian,

---

[5] We use "longer" here to mean having more outlinks in order to simplify the discussion.

Portuguese, Romanian, English, German, French, Italian, and Danish.

To validate our implementation, we tested our English version against a canonical human gold standard dataset in semantic relatedness (Miller and Charles 30 [19]) and achieved correlations that were comparable to the original ESA implementations ($r_s = 0.71$). Since in our experiment we would be comparing between many named entity pairs, we also evaluated our implementation against the only named entity SR dataset available (to our knowledge) [24], which comes from the bioinformatics domain. Interestingly, our English ESA ($r_s = 0.79$) performed comparably to Pedersen et al.'s [24] SR measure ($r_s = 0.84$) even though Pedersen et al.'s was based on large amounts of expert knowledge.

ESA is optimal as it allows us to change the concepts that are used for modeling. In other words, we can plug in world knowledge that varies in language and content relatively easily. Using this flexibility, we performed two experiments with ESA. In the first, we used as world knowledge the 8,264 perfect clarity concepts (with |inlinks| and |outlinks| >= 3) that existed in the intersection between the 10 languages in the study (or rather, the specific article instances of these concepts in each language). This models the pure effect of sub-concept diversity on ESA without any concept-level effect. If $C_1$ = "Argentina" and $C_2$ = "Sigmund Freud", how much will $ESA_{SPANISH}(C_1, C_2)$ differ from $ESA_{GERMAN}(C_1, C_2)$ due to the sub-concept diversity in the "Psychology" article?

In the second experiment, we used as world knowledge 10,000 randomly selected articles from each language (each article was required to have five outlinks to ensure sufficient content). This is done to model another way in which ESA is typically implemented and because it includes concept-level diversity as well as sub-concept level diversity. To understand the possible effect of concept-level diversity on ESA results, consider $C_1$ = "Country Music" and $C_2$ = "Duo". If the "Big and Rich" article appeared in the $ESA_{ENGLISH}$ 10,000 concepts, $ESA_{ENGLISH}$ would understand a relationship between $C_1$ and $C_2$ that none of the other ESAs would be able to understand, increasing the resultant SR value relative to the other ESA values.

For both experiments, we tested on 2,000 $C_1,C_2$ concept pairs randomly selected from the list of global concepts discussed above. For each language, $C_1$ and $C_2$ were set to the title of the article of the concept in each language. We only used concepts that have single-word titles in all ten languages. Two pairs where $C_1 = C_2$ were included, and this is in line with typical SR human gold standard datasets such as WordSim353 [10].

**Results**

The effect of sub-concept diversity on the output values of ESA was extensive. The mean correlation coefficient $r$ was only 0.13 (0.16 without $C_1 = C_2$ pairs). In other words, ESA generates very different semantic relatedness values depending on the culture whose world knowledge is used. Had this not been the case, the correlations would have been much closer to 1.0.

|          | Spanish | English | German | French |
|----------|---------|---------|--------|--------|
| **English**  | 0.16    | 1.00    | 0.11   | 0.09   |
| **French**   | 0.17    | 0.09    | 0.09   | 1.00   |
| **German**   | 0.11    | 0.11    | 1.00   | 0.09   |
| **Romanian** | 0.33    | 0.14    | 0.16   | 0.15   |

Table 5: Correlation coefficients between SR values generated by ESA systems based on different languages. This is a subset of the 10x10 matrix generated by the study.

Because of the differentiation in content, pairs that may appear very related according to one ESA may not be considered related according to another. Consider, for example, the concept pair "Germany" / "Saxony-Anhalt". In most of the languages, this pair receives high ESA scores, but in Italian and Danish, ESA perceives no relation at all. This is because in Italian and Danish, the articles that make up the world knowledge do not mention "Germany" and "Saxony-Anhalt" together (or do not mention one or both at all) whereas the other languages do. Likewise, the "Triumphal Arch" article in English mentions both an arch in Thessaloniki and in Iraq, whereas the German article only mentions the Thessaloniki arch. Since "Triumphal Arch" is one of the ten-language concepts, this leads to "Thessaloniki" / "Iraq" having a much higher SR value in English than in German. That is not to say there was not widespread "agreement" amongst all ESAs on certain concept pairs. For instance, when $C_1$ = "1945" and $C_2$ = "1947", all the ESAs returned relatively high values. Obviously, these words occur frequently together in Wikipedia articles, regardless of the language. Similarly, many pairs such as "DVD" / "Djibouti" are not related in any language.

The results of the second experiment show similar trends to the first: mean $r = 0.16$ (0.18 without $C_1 = C_2$ pairs). In this experiment, there are two forces behind the low correlations: concept differences and language differences. To tease out the effect of the language difference, we compared the values from the multilingual experiment to those from an identical experiment using 16 random English-only 10,000-article sets as world knowledge. The mean of the English-only simulation ($r_{ENG16}$ =.77 or .42 without the $C_1 = C_2$ pairs) was significantly higher ($p < 0.001$) than the mean $r$ in the multilingual experiment ($r_{MULT}$ = .16 or .18 without the $C_1 = C_2$ pairs). In other words, given two ESA implementations, no matter which two 10,000-article sets are used as world knowledge, if those sets are from the same language they will lead to more similar SR scores than if the sets are from different languages (on average).

**DISCUSSION**

Throughout this paper we aimed to examine the veracity of the global consensus hypothesis, quantify the degree of knowledge diversity present across various Wikipedia language editions, and demonstrate the influence this knowledge diversity may have on technology.

For researchers in HCI, AI and NLP, the rejection of the global consensus hypothesis has important implications for technologies that operate on Wikipedia directly. At the concept level, this work places an important boundary condition on the utility of ideas like "information arbitrage" [2], which seeks to "leverage articles in one or more languages to improve content in another". For instance, a running example in [2] is articles on the concept "Jerry Seinfeld", a concept that does not exist in eight of the languages explored in this study. It is likely that for at least some of these eight language groups, Jerry Seinfeld has insufficient cultural import to warrant and maintain an article. More generally, it is possible that information arbitrage would be of little utility for a portion of the Wikipedia articles that only exist in a single language.

Ideas such as information arbitrage are also affected by sub-concept diversity. The wide-ranging extent of sub-concept diversity captures a great deal of culture-specific content and researchers must be aware that sub-concept diversity does not simply represent information "inefficiencies" that need to be fixed. Propagating culture-specific information such as that found in the "*Psicología*" article to other Wikipedia language editions, for example, would likely be detrimental to end-users. Readers of the Danish Wikipedia likely would not consider a section on psychology in Latin America to be very relevant in the Danish "*Psykologi*" article. The research challenge ahead is learning to automatically separate culture-specific information – such as geographically focused examples – from information that is largely globally relevant, like dates of birth, etc. This challenge plays an important role in future work.

The lack of global consensus also has large indirect effects on the larger class of Wikipedia-based applications. These are applications that use Wikipedia as a source of world knowledge to do non-Wikipedia actions. This directly impacts end-user applications that implement ESA. For example, Bergstrom and Karahalios's recent research [3] on the clustering of conversation topics on a shared display relied upon an English Wikipedia-based ESA metric. However, if the group conversing is a multinational team of scientists, our results suggest that clustering will be biased toward any native English speakers in the group, as it would be their world knowledge used for the clustering. The results for a conversation that involved Thessaloniki and Iraq would be different than if the German Wikipedia were used.

**Implications for Design**
While knowledge diversity can create problems in existing technologies, it opens up opportunities for developing new approaches to technology design. In particular, the potential for *culturally-aware applications* is enormous.

A culturally-aware application is an application that can swap between different representations of world knowledge as context demands. For instance, if a system needs to calculate the semantic relatedness between two entities for a group of Romanian immigrants to the United States, the Romanian Wikipedia could be swapped in and the English Wikipedia swapped out.

Another important and exciting subset of culturally-aware applications is *hyperlingual* applications, which consider world knowledge from multiple languages simultaneously. For instance, the work of Bergstrom and Karahalios could be extended hyperlingually if a weighted combination of the native languages of the participants were considered.

A hyperlingual approach can provide enormous benefits in terms of access to new world knowledge not available in any particular language edition (including English). Researchers have not yet taken advantage of the articles lying outside the concept intersection between languages, despite the fact that these articles far outnumber articles in the intersection. In other words, by donning a hyperlingual lens, technologists can utilize the diversity in knowledge representations as well as the similarities.

**Future Directions**
We are currently working on two hyperlingual applications that leverage the concept and sub-concept diversity findings in this paper. The first is effectively a "cultural reading level" application that will help people writing for an international audience to identify parochial or region-specific references, as well as suggest appropriate alternatives. Secondly, we are building a system to help foster the understanding of concept and sub-concept diversity in Wikipedia. The system will allow users to view the intersection and union of world knowledge on a particular concept. Importantly, it will also highlight the difference between the union and intersection for the user. We used a text-only nascent implementation of this system to identify some of the examples in this paper.

**CONCLUSION**
In this paper, we have provided four key contributions: (1) we have shown that knowledge diversity across Wikipedias is large and defined its extent, (2) we have demonstrated that this diversity has a significant effect on technologies, (3) the first census of the effect of language on UGC repositories was executed, and (4) we have discussed design implications of these findings while introducing the ideas of culturally-aware applications and hyperlingual applications. Moving forward, we hope this work will inform and inspire a new generation of multilingual Wikipedia applications.

**ACKNOWLEDGEMENTS**
We would like to thank Dr. Martin Raubal for his valuable insight and our labmates for their feedback. This work was supported in part by National Science Foundation grant #0705901 and the Robert and Kaye Hiatt fund.